\documentclass[11pt,a4paper]{article}
\usepackage[hyperref]{acl2021}
\usepackage{times}
\usepackage{latexsym}

\usepackage{makecell}
\usepackage{bm}
\usepackage{amssymb}
\usepackage{microtype}
\usepackage{graphicx} 
\usepackage{float} 
\usepackage{hyperref}
\usepackage{url}
\usepackage{booktabs}
\usepackage{multirow}
\usepackage{graphics}
\usepackage{amsmath}
\usepackage{multicol}
\usepackage{lipsum}
\usepackage{mwe}
\usepackage{hyperref}
\usepackage{url}
\usepackage{subfig}
\usepackage{xcolor}
\usepackage{bbm}
\usepackage{upgreek}

\aclfinalcopy % Uncomment this line for the final submission

\def\aclpaperid{2993} %  Enter the acl Paper ID here

\newcommand{\sectionref}[1]{\S\ref{#1}}

\title{AdaTag: Multi-Attribute Value Extraction from Product Profiles\\with Adaptive Decoding}

\author{Jun Yan$^{1}$\thanks{~~Most of the work was done during an internship at Amazon.}~~, Nasser Zalmout$^{2}$, Yan Liang$^{2}$, Christan Grant$^{3}$, 
\\\textbf{Xiang Ren$^{1}$, Xin Luna Dong$^{2}$} \\
University of Southern California$^{1}$, Amazon.com$^{2}$, University of Oklahoma$^{3}$, \\
\small{\texttt{\{yanjun, xiangren\}@usc.edu}}, \small{\texttt{\{nzalmout, ynliang, lunadong\}@amazon.com}}, \small{\texttt{cgrant@ou.edu}}
}

\begin{document}
\maketitle
\begin{abstract}
%% Xiang
Automatic extraction of product attribute values is an important enabling technology in e-Commerce platforms.
This task is usually modeled using sequence labeling architectures, with several extensions to handle multi-attribute extraction.
One line of previous work constructs attribute-specific models, through separate decoders or entirely separate models. However, this approach constrains knowledge sharing across different attributes. Other contributions use a single multi-attribute model, with different techniques to embed attribute information. But sharing the entire network parameters across all attributes can limit the model's capacity to capture attribute-specific characteristics.
% Previous contributions use separate networks for each attribute, whether through separate decoders or totally separate models, which is not scalable and constrains knowledge sharing across different attributes.
% Others contributions use a single model to handle all attributes, resulting in a limited capacity to capture attribute-specific characteristics.
In this paper we present AdaTag, which uses adaptive decoding to handle extraction. We parameterize the decoder with pretrained attribute embeddings, through a hypernetwork and a Mixture-of-Experts (MoE) module. This allows for separate, but semantically correlated, decoders to be generated on the fly for different attributes. This approach facilitates knowledge sharing, while maintaining the specificity of each attribute.
Our experiments on a real-world e-Commerce dataset show marked improvements over previous methods.

\end{abstract}
\section{Introduction}

\begin{figure}[t]
\centering
\includegraphics[scale=0.41]{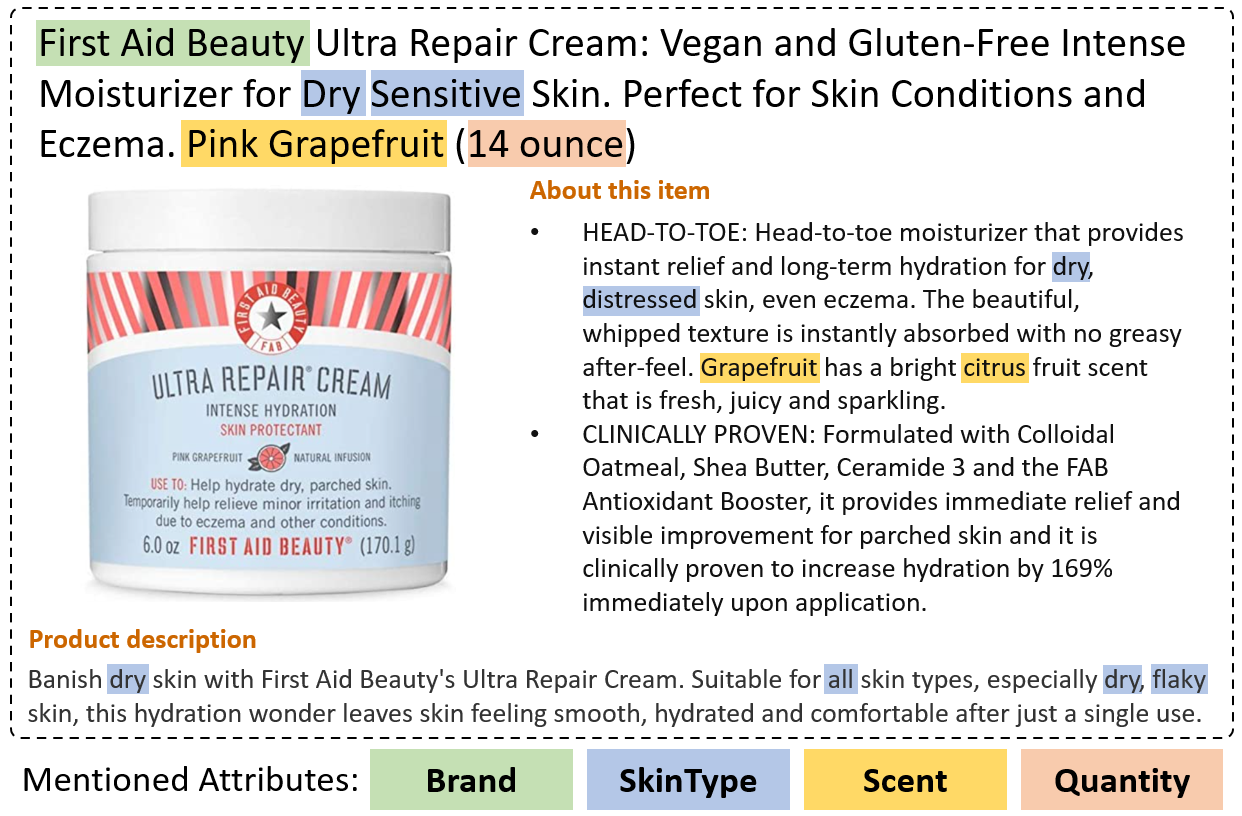}
\caption{An example of the product profile on an e-Commerce platform. It consists of a title, several information bullets, and a product description.}
% \vspace{-2mm}
\label{fig:example}
\end{figure}

The product profiles on e-Commerce platforms are usually comprised of natural texts describing products and their main features. Key product features  are conveyed in unstructured texts, with limited impact on machine-actionable applications, like search \citep{ai2017learning}, recommendation \citep{kula2015metadata}, and question answering \citep{kulkarni2019productqna}, among others. Automatic attribute value extraction aims to obtain structured product features from product profiles. The input is a textual sequence from the product profile, along with the required attribute to be extracted, out of potentially large number of attributes. The output is the corresponding extracted attribute values. 
Figure~\ref{fig:example} shows the profile of a moisturizing cream product as an example, which consists of a title, several information bullets, and a product description. It also shows the attribute values that could be extracted.

% We aim at extracting the words ``\textit{dry}'' and ``\textit{sensitive}'' as values for the ``\textit{SkinType}'' attribute, for example. Having access to the other attributes while training can also help the model in disambiguating the value ``\textit{dry}'' as ``\textit{SkinType}'', and avoid confusing it with ``\textit{HairType}'', for instance.

Most existing studies on attribute value extraction use neural sequence labeling architectures \citep{zheng2018opentag, karamanolakis2020txtract, xu2019scaling}. To handle multiple attributes, one line of previous contributions develops a set of ``attribute-specific" models (i.e., one model per attribute). The goal is to construct neural networks with (partially) separate model parameters for different attributes. For example, one can construct an independent sequence labeling model for each attribute and make predictions with all the models collectively (e.g., the vanilla OpenTag model \citep{zheng2018opentag}). Instead of totally separate models, one can also use different tag sets corresponding to different attributes. These networks can also share the feature encoder and use separate label decoders \citep{DBLP:conf/iclr/YangSC17}. However, the explicit network (component) separation in these contributions constrains knowledge-sharing across different attributes. Exposure to other attributes can help in disambiguating the values for each attribute. And having access to the entire training data for all attributes helps with the generic sequence tagging task.
%Moreover, the number of networks in these approaches would scale linearly with the number of attributes. This puts these models at a disadvantage when scaling to a large number of attributes \citep{xu2019scaling}.
%
Another line for multi-attribute extraction contributions learns a \textit{single} model for all attributes. The model proposed by \citet{xu2019scaling}, for example, embeds the attribute name with the textual sequence, to achieve a single ``attribute-aware'' extraction model for all attributes. This approach addresses the issues in the previous direction. However, sharing all the network parameters with all attributes could limit the model's capacity to capture attribute-specific characteristics. 

In this paper we address the limitations of the existing contribution lines, through \textit{adaptive decoder parameterization}. We propose to generate a decoder on the fly for each attribute based on its embedding. This results in different but semantically correlated decoders, which maintain the specific characteristics for each attribute, while facilitating knowledge-sharing across different attributes. To this end, we use conditional random fields (CRF) \citep{lafferty2001conditional} as the decoders, and parameterize the 
decoding layers with the attribute embedding through a hypernetwork \citep{ha2016hypernetworks} and a Mixture-of-Experts (MoE) module \citep{jacobs1991adaptive}.
%%%
We further explore several pretrained attribute embedding techniques, to add useful attribute-specific external signals. We use both contextualized and static embeddings for the attribute name along with its potential values to capture meaningful semantic representations.

% In this paper we address the limitations of both lines of existing contributions, through adaptive decoding based on conditional random fields (CRF) \citep{lafferty2001conditional}. We share the encoder across all attributes, but the parameters of the decoder are generated on-the-fly based on the target attribute. We parameterize the CRF-based decoder with the embedding of the attribute, through a hypernetwork \citep{ha2016hypernetworks} and a Mixture-of-Experts (MoE) \citep{jacobs1991adaptive} module. This setup results in different but semantically correlated decoders for the different attributes, based on the attribute embedding. This provides the model with the capacity to capture attribute-specific characteristics, while facilitating knowledge-sharing across the attributes. This model is also more scalable to larger number of attributes, since the decoders are dynamically created, as opposed to using static separate decoders. We further explore pretrained attribute embeddings, to add useful attribute-specific external signals. We use both contextualized and static embedding methods, using the attribute name along with a sample of the attribute's values, to capture more meaningful semantic representation.

We summarize our contributions as follows: (1) We propose a multi-attribute value extraction model with an adaptive CRF-based decoder. Our model allows for knowledge sharing across different attributes, yet maintains the individual characteristics of each attribute. (2) We propose several attribute embedding methods, that provide important external semantic signals to the model. (3) We conduct extensive experiments on a real-world e-Commerce dataset, and show improvements over previous methods. We also draw insights on the behavior of the model and the attribute value extraction task itself.

\section{Background}

\subsection{Problem Definition}

The main goal of the task is to extract the corresponding values for a given attribute, out of a number of attributes of interest, from the text sequence of a product profile.
Formally, given a text sequence $X=[x_1, \ldots, x_n]$ in a product profile, where $n$ is the number of words, and a query attribute $r\in{R}$, where $R$ is a predefined set of attributes, the model is expected to extract all text spans from $X$ that could be valid values for attribute $r$ characterizing this product.
When there are no corresponding values mentioned in $X$, the model should return an empty set.
For example, for the product in Figure~\ref{fig:example}, given its title as $X$, the model is expected to return (``Dry'', ``Sensitive'') if $r=$``SkinType'', and an empty set if $r=$``Color''.

Following standard approaches \citep{zheng2018opentag, xu2019scaling, karamanolakis2020txtract}, under the assumption that different values for an attribute do not overlap in the text sequence, we formulate the value extraction task as a sequence tagging task with the BIOE tagging scheme.
That is, given $X$ and $r$, we want to predict a tag sequence $Y=[y_1, \ldots, y_n]$, where $y_i\in\{\text{B},\text{I},\text{O},\text{E}\}$ is the tag for $x_i$.
``B''/``E'' indicates the corresponding word is the beginning/ending of an attribute value, ``I'' means the word is inside an attribute value, and ``O'' means the word is outside any attribute value.
Table~\ref{tab:bioe} shows an example of the tag sequence for attribute ``Scent'' of a shower gel collection, where ``orchid'', ``cherry pie'', ``mango ice cream'' could be extracted as the values.
\begin{table}[h]
% \vspace{-0.1cm}
\centering
\scalebox{0.65}{
\begin{tabular}{c|ccccccccc}
    \toprule
    % X & duck & , & filet & mignon & and & ranch & raised & lamb & flavor \\
    % \midrule
    % Y & B & O & B & E & O & B & I & E & O \\
    X & orchid & / & cherry & pie & / & mango & ice & cream & scent \\
    \midrule
    Y & B & O & B & E & O & B & I & E & O \\
    \bottomrule
\end{tabular}
}
% \vspace{-0.2cm}
\caption{An example of the tag sequence for attribute ``Scent'' annotated with the BIOE scheme.}
\label{tab:bioe}
% \vspace{-0.3cm}
\end{table}

\subsection{BiLSTM-CRF Architecture}
\label{sec:bilstm_crf}
The BiLSTM-CRF architecture \citep{huang2015bidirectional} consists of a BiLSTM-based text encoder, and a CRF-based decoder. This architecture has been proven to be effective for the attribute value extraction task \citep{zheng2018opentag, xu2019scaling, karamanolakis2020txtract}.
We build our AdaTag model based on the BiLSTM-CRF architecture as we find that the BiLSTM-CRF-based models generally perform better than their BiLSTM-based, BERT-based~\citep{devlin2018bert} and BERT-CRF-based counterparts, as shown in \sectionref{sec:results}.
We introduce the general attribute-agnostic BiLSTM-CRF architecture, which our model is based on, in this subsection.

Given a text sequence $X=[x_1, \ldots, x_n]$.
We obtain the sequence of word embeddings $\mathbf{X}=[\mathbf{x}_1, \ldots, \mathbf{x}_n]$ using an embedding matrix $\mathbf{W}_\text{word}$.
We get the hidden representation of each word by feeding $\mathbf{X}$ into a bi-directional Long-Short Term Memory (LSTM) \citep{hochreiter1997long} layer with hidden size $d_h$:
\begin{equation}
[\mathbf{h}_1, \ldots, \mathbf{h}_n] = \texttt{BiLSTM}([\mathbf{x}_1, \ldots, \mathbf{x}_n]).
\end{equation}
We use a CRF-based decoder to decode the sequence of hidden representations while capturing the dependency among tags (e.g., ``I'' can only be followed by ``E'').
It consists of a linear layer and a transition matrix, which are used to calculate the emission score and the transition score for the tag prediction respectively.
Let $V=[\text{B}, \text{I}, \text{O}, \text{E}]$ be the vocabulary of all possible tags.
We calculate an emission score matrix $\mathbf{P}=[\mathbf{p}_1,\ldots,\mathbf{p}_n]\in \mathbb{R}^{4\times n}$, where $\mathbf{P}_{ij}$ is the score for assigning the $i$-th tag in $V$ to $x_j$.
This is computed by feeding $[\mathbf{h}_1, \ldots, \mathbf{h}_n]$ into a linear layer with parameters $[\mathbf{W}, \mathbf{b}]$, specifically $\mathbf{p}_i=\mathbf{Wh}_i+\mathbf{b}\in\mathbb{R}^4$, where $\mathbf{W}\in \mathbb{R}^{4\times d_h}$ and $\mathbf{b}\in\mathbb{R}^4$.
For a BIOE tag sequence $Y=[y_1, \ldots, y_n]$, we get its index sequence $Z=[z_1, \ldots, z_n]$ where $z_i\in\{1,2,3,4\}$ is the index of $y_i$ in $V$.
The score for an input text sequence $X$ to be assigned with a tag sequence $Y$ is calculated as:
\begin{equation}
\label{eq:likelihood}
s(X,Y)=s(X,Z)=\sum_{i=1}^{n-1} \mathbf{T}_{z_i z_{i+1}} + \sum_{i=1}^n \mathbf{P}_{z_i i},
\end{equation}
where $\mathbf{T}\in\mathbb{R}^{4\times 4}$ is the transition matrix of CRF, such that $\mathbf{T}_{ij}$ is the score of a transition from the $i$-th tag to the $j$-th tag in $V$.

\section{Method}
\subsection{Model Overview}
\begin{figure*}[ht]
% \vspace{-0.1cm}
\centering
\includegraphics[scale=0.45]{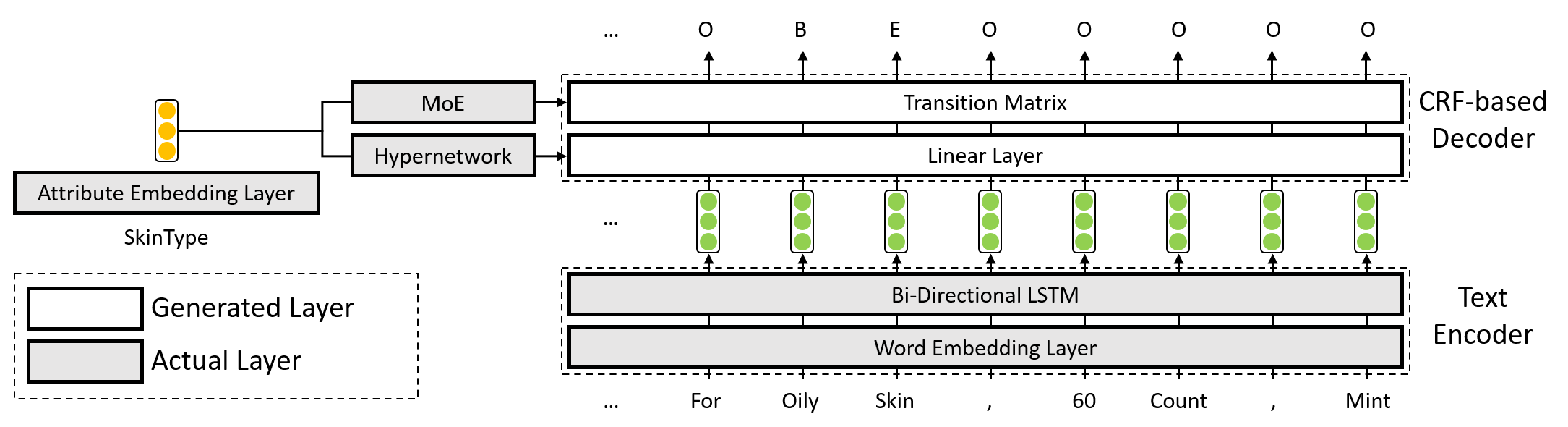}
% \vspace{-0.3cm}
\caption{Model architecture. AdaTag equips the BiLSTM-CRF architecture with an adaptive CRF-based decoder.}
\label{fig:architecture}
% \vspace{-0.2cm}
\end{figure*}

The multi-attribute value extraction task can be thought of as a group of extraction subtasks, corresponding to different attributes.
While all attributes share the general knowledge about value extraction, each has its specificity.
The key idea in our proposed model is to dynamically adapt the parameters of the extraction model based on the specific subtask corresponding to the given attribute. 
We use a BiLSTM-CRF \citep{huang2015bidirectional} architecture, where different subtasks, corresponding to different attributes, share the same text encoder to derive a contextualized hidden representation for each word.
Then the hidden representations of the text sequence are decoded into a sequence of tags with a CRF-based decoder, the parameters of which are generated on the fly based on the attribute embedding. 
In this setup, different subtasks are trained jointly, and different decoders are correlated based on the attribute embedding. This facilitates a knowledge-sharing scheme across different attributes.
Intuitively, this can help with learning generic abilities like detecting value boundary, which is at the core of the extraction process of any attribute.
%This is also beneficial due to the semantic similarities between the attributes.
At the same time, our model provides each subtask with a customized decoder parameterization, which improves the model's capacity for capturing attribute-specific knowledge.

Figure~\ref{fig:architecture} presents our overall model architecture, where we equip the BiLSTM-CRF architecture with an adaptive CRF-based decoder. In \sectionref{sec:adaptive}, we will introduce our adaptive CRF-based decoder which is parameterized with the attribute embedding. In \sectionref{sec:att_emb}, we will describe how to obtain pretrained attribute embeddings that can capture the characteristics of different subtasks, so that ``similar'' attributes get ``similar'' decoding layers.

\subsection{Adaptive CRF-based Decoder}
\label{sec:adaptive}

% \xiang{Can we have an illusrative figure for the ideas in this part?}

In attribute value extraction, the model takes the text sequence $X$ with a query attribute $r$ as input, and is expected to predict $Y$ based on both $X$ and $r$.
To make the model aware of the query attribute, we need to incorporate the attribute information into some components of the BiLSTM-CRF architecture.
The BiLSTM-based text encoder is responsible for encoding the text sequence and obtain a contextualized representation for each word, which can be regarded as ``understanding'' the sentence.
The CRF-based decoder then predicts a tag for each word based on its representation.
Therefore, we propose that all attributes share a unified text encoder so that the representation can be enhanced through learning with different subtasks, and each attribute has a decoder adapted to its corresponding subtask, the parameters of which are generated based on the attribute information.

As introduced in \sectionref{sec:bilstm_crf}, a CRF-based decoder consists of a linear layer and a transition matrix.
The linear layer takes hidden representations as input, and predicts a tag distribution for each word independently. It captures most of characteristics of value extraction for a given attribute based on the text understanding.
More flexibility is needed to model the specificity of different attributes.
By contrast, the transition matrix learns the dependency among tags to avoid predicting unlikely tag sequence. It only captures shallow characteristics for the attribute based on its value statistics.
For example, the transition scores form ``B'' to other tags largely depend on the frequent lengths of the attribute values.
If single-word values are mentioned more often, then ``B'' is more likely to be followed by ``O''.
If two-word values dominate the vocabulary, then ``B'' is more likely to be followed by ``E''.
Attributes could be simply clustered based on these shallow characteristics.

In this work we parameterize the CRF-based decoder with the attribute embedding $\mathbf{r}\in\mathbb{R}^{d_r}$, where $d_r$ is the dimension of the attribute embedding. For the linear layer, we adopt a hypernetwork \citep{ha2016hypernetworks} due to its high flexibility. For the transition matrix, we develop a Mixture-of-Experts \citep{pahuja2019learning} module to leverage the latent clustering nature of attributes. 
We nevertheless experiment with all 4 combinations of these methods in \sectionref{sec:ablation}, and this choice does the best.

% adopt the ideas of hypernetworks \citep{ha2016hypernetworks}, and Mixture-of-Experts \citep{pahuja2019learning} to parameterize the decoder layers with the attribute embedding $\mathbf{r}\in\mathbb{R}^{d_r}$, where $d_r$ is the dimension of the attribute embedding. Hypernetworks convey a certain level of flexibility, which makes them suitable to parameterize  $[\mathbf{W}, \mathbf{b}]$ of the linear layer. Whereas Mixture-of-Experts is more capable of capturing the relationships between different attributes, so we use it to parameterize the transition matrix $\mathbf{T}$ in CRF, which tends to have a more structured pattern for the predictions. 

\paragraph{Hypernetwork.}
The idea of hypernetworks \citep{ha2016hypernetworks} is to use one network to generate the parameters of another network. Such approach has high flexibility when no constraint is imposed during generation. We therefore use it to parameterize the linear layer. In our model, we learn two different linear transformations that map the attribute embedding to the parameters of the linear layer ($\mathbf{W}\in\mathbb{R}^{4\times d_h}$, $\mathbf{b}\in\mathbb{R}^4$) in the CRF-based decoder:
\begin{equation}
\begin{split}
\mathbf{W}&=\texttt{Reshape}(\mathbf{W}_\text{hyper}^w \mathbf{r}+\mathbf{b}_\text{hyper}^w),\\
\mathbf{b}&=\texttt{Reshape}(\mathbf{W}_\text{hyper}^b \mathbf{r}+\mathbf{b}_\text{hyper}^b).
\end{split}
\end{equation}
Here $\mathbf{W}_\text{hyper}^w\in\mathbb{R}^{4d_h\times d_r}$, $\mathbf{b}_\text{hyper}^w\in\mathbb{R}^{4d_h}$, $\mathbf{W}_\text{hyper}^b\in\mathbb{R}^{4\times d_r}$, $\mathbf{b}_\text{hyper}^b\in\mathbb{R}^{4}$, and the $\texttt{Reshape}$ operator reshapes a $1$-D vector into a matrix with the same number of elements.

\paragraph{Mixture-of-Experts.}
The idea of Mixture-of-Experts \citep{jacobs1991adaptive} is to have a group of networks (``experts'') that jointly make decisions with dynamically determined weights. Unlike previous approaches that combine each expert's \textit{prediction}, we combine their \textit{parameters} for generating the transition matrix.
% Some attributes share more characteristics than others.
% For example, SkinType and HairType have many values (\textit{e.g.} dry, oily) in common, the values for CoffeeRoastType and CaffeineContent have similar mentioning context.
% Intuitively, it should be beneficial for them to have more shared weights that can help joint modeling.
% Motivated by this, we adopt a Mixture-of-Experts (MoE) \citep{jacobs1991adaptive} design to detect latent clusters among all attributes and represent the parameters for a given attribute as a weighted combination of some basis parameters.
Let $k$ be the number of experts we use to parameterize the transition matrix $\mathbf{T}\in\mathbb{R}^{4\times 4}$ where $k$ is a hyperparameter.
We introduce $k$ learnable matrices $\mathbf{T}^{(1)}, \ldots, \mathbf{T}^{(k)}$ for the $k$ experts.
Each expert's matrix can be understood as a cluster prototype and we employ a linear gating network to compute the probability of assigning the given attribute to each expert:
$
\bm{\lambda}=\texttt{Softmax}(\mathbf{W}_\text{moe}\mathbf{r}+\mathbf{b}_\text{moe})
$.
Here $\mathbf{W}_\text{moe}\in\mathbb{R}^{k\times d_r}$, $\mathbf{b}_\text{moe}\in\mathbb{R}^{k}$, $\bm{\lambda}=[\lambda_1, \ldots, \lambda_k]\in\mathbb{R}^{k}$ and $\sum_{i=1}^k \lambda_i=1$.
The parameters for the transition matrix for this attribute is calculated as:
$
\mathbf{T} = \sum_{i=1}^k \lambda_i\mathbf{T}^{(i)}
$.

\subsection{Pretrained Attribute Embeddings}
\label{sec:att_emb}

%\xiang{Can we have an illusrative figure for the ideas in this part?}

The attribute embedding $\mathbf{r}$ plays a key role in deriving the attribute-specific decoding layers.
Therefore, the quality of the attribute embeddings is crucial to the success of our parameterization method.
Good attribute embeddings are supposed to capture the subtask similarities such that similar extraction tasks use decoders with similar parameters.
In this work, we propose to use the attribute name and possible values as a proxy to capture the characteristics of the value extraction task for a given attribute.
The attribute embeddings can therefore be directly derived from the training data and loaded into the attribute embedding layer as initialization.

For each attribute $r$, we first collect all the sentences from the training data that are annotated with at least one value for $r$.
We denote the collected sentences with values as $D_r=\{(\tilde{r}, v_i, X_i)\}_{i=1}^{n_r}$ where $\tilde{r}$ is the phrase representation of $r$ (e.g., $\tilde{r}=$ ``Skin Type'' if $r=$ ``SkinType''), $v_i$ is a span in text sequence $X_i$ that serves as the value for $r$, and $n_r$ is the number of collected sentences.
For each $(\tilde{r}, v_i, X_i)$, we can calculate an attribute name embedding $\mathbf{r}_i^{\text{name}}$ and an attribute value embedding $\mathbf{r}_i^{\text{value}}$ in either a contextualized way or an uncontextualized way, which are detailed later.
We pool over all instances in $D_r$ to get the final attribute name embedding and attribute value embedding, which are concatenated as the attribute embedding: 
$\mathbf{r}^{\text{name}} = \frac{1}{n_r}\sum_{i=1}^{n_r}\mathbf{r}_i^{\text{name}}$,
$\mathbf{r}^{\text{value}} = \frac{1}{n_r}\sum_{i=1}^{n_r}\mathbf{r}_i^{\text{value}}$,
$\mathbf{r}=\texttt{Concat}(\mathbf{r}^{\text{name}}, \mathbf{r}^{\text{value}})$.
% \begin{equation}
% \begin{split}
% &\mathbf{r}^{\text{name}} = \frac{1}{n_r}\sum_{i=1}^{n_r}\mathbf{r}_i^{\text{name}}
% \quad\mathbf{r}^{\text{value}} = \frac{1}{n_r}\sum_{i=1}^{n_r}\mathbf{r}_i^{\text{value}},\\
% &\mathbf{r}=\texttt{Concat}(\mathbf{r}^{\text{name}}, \mathbf{r}^{\text{value}}).
% \end{split}
% \end{equation}

\paragraph{Contextualized Embeddings.}
Taking the context into consideration helps get embeddings that can more accurately represent the semantics of the word. Here we use the contextualized representations provided by BERT \citep{devlin2018bert} to generate the embedding. We use BERT to encode $X_i$ and get $v_i$'s phrase embedding (the averaged embedding of each word in the phrase) as $\mathbf{r}_i^{\text{value}}$. By replacing $v_i$ with ``[BOA] $\tilde{r}$ [EOA]''\footnote{[BOA] and [EOA] are special tokens that are used to separate the attribute name from context in the synthetic sentence.} and encoding the modified sequence with BERT, we get the phrase embedding for ``[BOA] $\tilde{r}$ [EOA]'' as $\mathbf{r}_i^{\text{name}}$.

\paragraph{Uncontextualized Embeddings.}
Static embeddings like Word2Vec \citep{mikolov2013efficient} and Glove \citep{pennington2014glove} can be more stable to use under noisy contexts. We use Glove (50d) to get the phrase embedding for $v_i$ as $\mathbf{r}_i^{\text{value}}$ and the phrase embedding for $\tilde{r}$ as $\mathbf{r}_i^{\text{name}}$.

\subsection{Model Training}

As we parameterize the CRF-based decoder with the attribute embedding through MoE and hypernetwork, the learnable parameters in our model includes $\boldsymbol{\uptheta}_\text{encoder} = \{\mathbf{W}_\text{word},\boldsymbol{\uptheta}_\text{bi-lstm}\}$,
$\boldsymbol{\uptheta}_\text{hyper} = \{\mathbf{W}_\text{hyper}^w, \mathbf{b}_\text{hyper}^w, \mathbf{W}_\text{hyper}^b, \mathbf{b}_\text{hyper}^b\}$,
$\boldsymbol{\uptheta}_\text{moe} = \{\mathbf{W}_\text{moe},\mathbf{b}_\text{moe},\{\mathbf{T}^{(i)}\}_{i=1}^k\}$. We freeze the attribute embeddings $\mathbf{W}_\text{att}$ as it gives better performance, which is also discussed in \sectionref{sec:ablation}.

The whole model is trained end-to-end by maximizing the log likelihood of $(X,r, Y)$ triplets in the training set, which is derived from Equation~\ref{eq:likelihood} as:
\begin{equation}
\begin{split}
s(X,r,Y)&=\sum_{i=0}^n \mathbf{T}_{z_i z_{i+1}} + \sum_{i=1}^n \mathbf{P}_{z_i i},\\
\log p(Y\mid X,r)&=\log \frac{s(X,r,Y)}{\sum_{Y'\in V^n}s(X,r,Y')},
\end{split}
\end{equation}
where $V_n$ is the set of all tag sequences of length $n$.
The log likelihood can be computed efficiently using the forward algorithm \citep{baum1967inequality} for hidden Markov models (HMMs). At inference, we adopt Viterbi algorithm \citep{viterbi1967error} to get the most likely $Y$ given $X$ and $r$ in test set.

\section{Experimental Setup}
\subsection{Dataset}
\label{sec:data}

To evaluate the effectiveness of our proposed model, we build a dataset by collecting product profiles (title, bullets, and description) from the public web pages at \href{https://www.amazon.com/}{Amazon.com}.\footnote{While \citet{xu2019scaling} released a subset of their collected data from \href{https://www.aliexpress.com/}{AliExpress.com}, their data has a long-tailed attribute distribution (7650 of 8906 attributes occur less than 10 times). It brings major challenges for zero-/few-shot learning, which are beyond our scope.}

Following previous works \citep{zheng2018opentag, karamanolakis2020txtract, xu2019scaling}, we obtain the attribute-value pairs for each product using the product information on the webpages by distant supervision.
%We use a distant supervision approach for the data preparation as used in previous contributions \citep{zheng2018opentag, karamanolakis2020txtract}, using the product Catalog.\jun{I think this is not an informative description and could reveal that we're from Amazon.}
%For each product, the seller is asked to provide values for applicable attributes in the catalog.
%However, the annotations are often noisy as they are seller-generated and have not been manually validated by the platform.
%Therefore, they are regarded as distantly-supervised data and we use them to build the training and development sets.
We select 32 attributes with different frequencies.
For each attribute, we collect product profiles that are labeled with at least one value for this attribute.
We further split the collected data into training (90\%) and development (10\%) sets.

The annotations obtained by distant supervision are often noisy so they cannot be considered as gold labels.
To ensure the reliability of the evaluation results, we also manually annotated an additional testing set covering several attributes. We randomly selected 12 attributes from the 32 training attributes, took a random sample from the relevant product profiles for each attribute, and asked human annotators to annotate the corresponding values. %these 12 attributes in the sampled product profiles to build a clean test set for each attribute.
We ensured that there is no product overlapping between training/development sets and the test set.

Putting together the datasets built for each individual attribute, we end up with training and development sets for \textbf{32} attributes, covering \textbf{333,857} and \textbf{40,008} products respectively.
The test set has \textbf{12} attributes and covers \textbf{11,818} products.
Table~\ref{tab:dataset} presents the statistics of our collected dataset.
Table~\ref{tab:att_distribution} shows the attribute distribution of the training set.
% The statistics of our collected dataset with attribute distribution are presented in \sectionref{sec:att_dist}.
It clearly demonstrates the data imbalance issue of the real-world attribute value extraction data.
% It suggests that real-world data is very attribute-imbalanced, posing challenges to the joint modeling of attributes.

Most of the attribute values are usually covered in the title and bullets, since sellers would aim to highlight the product features early on in the product profile.
The description, on the other hand, can provide few new values complementing those mentioned in the title and bullets, but significantly increases the computational costs due to its length.
Therefore, we consider two settings for experiments: extracting from the title only (``\textbf{Title}'') and extracting from the concatenation of the title and bullets (``\textbf{Title + Bullets}'').

\begin{table}[th]
\centering
\scalebox{0.7}{
\begin{tabular}{ccrcc}
    \toprule
    \textbf{Split} & \textbf{\# Attributes} & \textbf{\# Products} & \thead{\textbf{Avg. \# Words}\\(Title)} & \thead{\textbf{Avg. \# Words}\\(Title+Bullets)}\\
    \midrule
    train & 32 & 333,857 & 20.9 & 113.4 \\
    dev & 32 & 40,008 & 21.0 & 113.7 \\
    test & 12 & 11,818 & 20.5 & 120.0 \\
    \bottomrule 
\end{tabular}
}
\caption{Statistics of our collected dataset.}
%\christan{Is the large increase in the Average number of words for Title+Bullets significant? Consider adding the std with the averages.}
\label{tab:dataset}
\end{table}

\begin{table}[th]
\centering
\scalebox{0.7}{
\begin{tabular}{ccl}
    \toprule
    \textbf{\# Products} & \textbf{\# Att.} & \textbf{Examples}\\
    \midrule
    $[10000, 50279]$ & $12$ & Color, Flavor, SkinType, HairType \\
    $[1000, 10000)$ & $10$ & ActiveIngredients, CaffeineContent \\
    $[100, 1000)$ & $6$ & SpecialIngredients, DosageForm \\
    $[15, 100)$ & $4$ & PatternType, ItemShape\\
    \bottomrule 
\end{tabular}
}
\caption{Frequencies of different attributes in the training set.}
\label{tab:att_distribution}
\end{table}

\subsection{Evaluation Metrics}

For each attribute, we calculate Precision/Recall/F$_1$ based on exact string matching.
That is, an extracted value is considered correct only if it exactly matches one of the ground truth values for the query attribute in the given text sequence.
We use Macro-Precision/Macro-Recall/Macro-F$_1$ (denoted as P/R/F$_1$) as the aggregated metrics to avoid bias towards high-resource attributes.
They are calculated by averaging per-attribute metrics.

\subsection{Compared Methods}

We compare our proposed model with a series of strong baselines for attribute value extraction.\footnote{We discuss the sizes of different models in Appendix \sectionref{sec:num_parameters}.}

\textbf{BiLSTM} uses a BiLSTM-based encoder. Each hidden representation is decoded independently into a tag with a linear layer followed by softmax.
\textbf{BiLSTM-CRF} \citep{huang2015bidirectional} uses a BiLSTM-based encoder and a CRF-based decoder, as described in \sectionref{sec:bilstm_crf}. \citet{zheng2018opentag} propose OpenTag, which uses a self-attention layer between the BiLSTM-based encoder and CRF-based decoder for interpretable attribute value extraction. However, we find the self-attention layer not helpful for the performance.\footnote{We hypothesize that the improvement brought by the self-attention module is dataset-specific.}
We therefore only present the results for BiLSTM-CRF in \sectionref{sec:results}.
\textbf{BERT} \citep{devlin2018bert} and \textbf{BERT-CRF} replace the BiLSTM-based text encoder with BERT.\footnote{The hidden representation for each word is the average of its subword representations.}

Note that these four methods don't take the query attribute as input.
To make them work in our more realistic setting with multiple ($N$) attributes, we consider two variants for each of them. (1) ``$N$ tag sets'': We introduce one set of B/I/E tags for each attribute, so that a tag sequence can be unambiguously mapped to the extraction results for multiple attributes. For example, the tag sequence ``B-SkinType E-SkinType O B-Scent'' indicates that the first two words constitutes a value for attribute SkinType, and the last word is a value for Scent. Only one model is needed to handle the extraction for all attributes. (2) ``$N$ models'': We build one value extraction model for each attribute --- we'll train $N$ models for this task. 

The ``$N$ models'' variant isolates the learning of different attributes. To enable knowledge sharing, other methods share the  model components or the whole model among all attributes:
\textbf{BiLSTM-CRF-SharedEmb} shares a word embedding layer among all attributes. Each attribute has its own BiLSTM layer and CRF-based decoder, which are independent from each other.
\textbf{BiLSTM-MultiCRF} \citep{DBLP:conf/iclr/YangSC17} shares a BiLSTM-based text encoder among all attributes. Each attribute has its own CRF-based decoder.
\textbf{SUOpenTag} \citep{xu2019scaling} encodes both the text sequence and the query attribute with BERT and adopts a cross-attention mechanism to get an attribute-aware representation for each word. The hidden representations are decoded into a tags with a CRF-based decoder.

We also include \textbf{AdaTag (Random AttEmb)}, which has the same architecture as our model but uses randomly initialized learnable attribute embeddings of the same dimension.
% These baselines can be categorized into 4 classes:
% (1) BiLSTM ($N$ models) and OpenTag ($N$ models) train $N$ models for the multi-attribute value extraction task, which makes the solution infeasible when facing a large number of attributes.
% There is also no interaction between the modeling of different attributes.
% (2) BiLSTM ($N$ tag sets), OpenTag (joint model) and MultiCRF develop one model for the task but the decoder design is not scalable with respect to the number of attributes due to either $N$ sets of B/I/E tags or $N$ sets of decoder parameters.

% (2) BiLSTM (joint model), OpenTag (joint model) and MultiCRF develop one model for the task but the decoder design is not scalable with respect to the number of attributes due to either $N$ sets of B/I/E tags or $N$ sets of decoder parameters.
% (4) SUOpenTag is the only model that has good scalability as our model. The number of parameters is independent of the number of attributes.

\subsection{Implementation Details}

We implement all models with PyTorch \citep{NEURIPS2019_9015}.
For models involving BERT, we use the \texttt{bert-base-cased} version.
Other models use pretrained 50d Glove  \citep{pennington2014glove} embeddings as the initialization of the word embedding matrix $\mathbf{W}_\text{word}$.
We choose $d_h=200$ as the hidden size of the BiLSTM layer and 32 as the batch size.
BERT-based models are optimized using AdamW \citep{loshchilov2017decoupled} optimizer with learning rate $2e^{-5}$. Others use the Adam \citep{kingma2014adam} optimizer with learning rate $1e^{-3}$.
We perform early stopping if no improvement in (Macro-) F$_1$ is observed on the development set for 3 epochs.
For our model, we use contextualized attribute embeddings as described in \sectionref{sec:adaptive} and freeze them during training.
We set $k=3$ for MoE.
We made choices based on the development set performance.

\section{Experimental Results}
\label{sec:results}
\subsection{Overall Results}

Table~\ref{tab:main} presents the overall results using our dataset under both ``Title'' and ``Title + Bullets'' settings.
Our model demonstrates great improvements over baselines on all metrics except getting second best recall under the ``Title + Bullets'' settings.
The comparisons demonstrate the overall effectiveness of our model and pretrained attribute embeddings.

\begin{table}[t]
\centering
\scalebox{0.60}{
\begin{tabular}{lcccccc}
    \toprule  
    \multirow{2}{*}{\textbf{Methods}}&
    \multicolumn{3}{c}{\textbf{Title}}&\multicolumn{3}{c}{ \textbf{Title + Bullets}}\\
    \cmidrule(lr){2-4} \cmidrule(lr){5-7}
     & P(\%) & R(\%) & F$_1$(\%) & P(\%) & R(\%) & F$_1$(\%) \\
    \toprule
    \multicolumn{7}{c}{Group I: $N$ tag sets}\\
    \midrule
    {BiLSTM ($N$ tag sets)} & 35.15 & 54.28 & 38.92 & 32.17 & 34.30 & 31.18 \\
    {BiLSTM-CRF ($N$ tag sets)} & 35.23 & 53.94 & 38.85 & 34.03 & 35.01 & 32.11 \\
    % {OpenTag ($N$ tag sets)} & 34.99 & 55.10 & 38.20 & 31.94 & 33.93 & 30.31 \\
    {BERT ($N$ tag sets)} & 33.52 & 50.48 & 36.29 & 31.41 & 30.62 & 28.26 \\
    {BERT-CRF ($N$ tag sets)} & 34.55 & 51.96 & 37.45 & 32.63 & 31.24 & 28.89 \\ 
    \toprule
    \multicolumn{7}{c}{Group II: $N$ models}\\
    \midrule  
    {BiLSTM ($N$ models)} & 64.37 & 71.71 & 64.64 & 61.61 & 60.26 & 58.56 \\
    {BiLSTM-CRF ($N$ models)} & 63.94 & 72.14 & 64.78 & \underline{62.07} & 61.46 & 59.19 \\ 
    % {OpenTag ($N$ models)} & 49.17 & 71.94 & 52.41 & 57.99 & 60.01 & 55.67 \\
    {BERT ($N$ models)} & 55.34 & \underline{72.86} & 58.48 & 53.35 & 61.27 & 54.37 \\
    {BERT-CRF ($N$ models)} & 54.29 & 72.79 & 57.49 & 49.25 & 59.33 & 50.49 \\
    \toprule
    \multicolumn{7}{c}{Group III: shared components}\\
    \midrule
    {BiLSTM-CRF-SharedEmb} & 63.77 & 72.50 & 64.62 & 58.95 & 60.58 & 57.66 \\ 
    {BiLSTM-MultiCRF} & 64.48 & 72.04 & 64.81 & 60.64 & \textbf{62.75} & \underline{59.78} \\
    {SUOpenTag} & 63.62 & 71.67 & 64.76 & 61.57 & 60.48 & 59.62 \\
    AdaTag (Random AttEmb) & \underline{64.80} & 71.95 & \underline{65.74} & 60.14 & 62.14 & 60.04  \\
    \textbf{AdaTag (Our Model)} & \textbf{65.00} & \textbf{75.87} & \textbf{67.48} & \textbf{62.87} & \underline{62.45} & \textbf{60.87} \\
    \bottomrule 
\end{tabular}
}
\caption{Performance comparison on test set with 12 attributes (best in boldface and second best underlined).}
\label{tab:main}
\end{table}

The ``$N$ tag sets'' variants get much lower performance than other methods, probably due to the severe data imbalance issue in the training set (see Table~\ref{tab:att_distribution}).
All attributes share the same CRF-based decoder, which could make learning biased towards high-resource attributes.
% todo: show results
Note that introducing one set of tags for each entity type is the standard approach for the Named Entity Recognition (NER) task.
Its low performance suggests that the attribute value extraction task is inherently different from standard NER.

Variants of ``shared components'' generally achieve higher performance than the independent modeling methods (``$N$ models''), which demonstrates the usefulness of enabling knowledge sharing among different subtasks.

We also notice that BERT and BERT-CRF models get lower performance than their BiLSTM and BiLSTM-CRF counterparts.
The reason could be the domain discrepancy between the corpora that BERT is pretrained on and the product title/bullets.
The former consist of mainly natural language sentences, while the latter are made up of integration of keywords and ungrammatical sentences.

\subsection{High- vs. Low-Resource Attributes}
To better understand the gain achieved by joint modeling, we further split the 12 testing attributes into 8 high-resource attributes and 4 low-resource attributes, based on the size of the training data with $1000$ instances as the threshold.
It is important to point out that many factors (e.g., vocabulary size, value ambiguity, and domain diversity), other than the size of training data, can contribute to the difficulty of modeling an attribute. Therefore, the performance for different attributes is not directly comparable.\footnote{Some low-resource attributes (e.g., BatteryCellComposition) have small value vocabulary and simple mentioning patterns. Saturated performance on them pull up the metrics.}
%we define high-resource attributes as attributes with more than $1000$ training samples, and low-resource attributes as attributes with fewer than $1000$ training samples. Out of 12 test attributes, 8 are high-resource and 4 are low-resource. Under ``Title'' setting, we calculate Macro-P/R/F$_1$ for high-resource test attributes and low-resource test attributes. 

From results in Table~\ref{tab:high_low_resource}, we can see that our model gets a lot more significant improvement from the independent modeling approach (BiLSTM-CRF ($N$ models)) on low-resource attributes compared to high-resource attributes.
This suggests that low-resource attributes benefit more from knowledge sharing, making our model desirable in the real-world setting with imbalanced attribute distribution.

\begin{table}[t]
\centering
\scalebox{0.60}{
\begin{tabular}{lcccccc}
    \toprule  
    \multirow{2}{*}{\textbf{Methods}}&
    \multicolumn{3}{c}{\textbf{High-Resource Att.}}&\multicolumn{3}{c}{ \textbf{Low-Resource Att.}}\\
    \cmidrule(lr){2-4} \cmidrule(lr){5-7}
     & P(\%) & R(\%) & F$_1$(\%) & P(\%) & R(\%) & F$_1$(\%) \\
    \midrule
    % {BiLSTM-CRF ($N$ tag sets)} & 25.98 & 53.96 & 32.51 & 53.72 & 53.90 & 51.53 \\ 
    {BiLSTM-CRF ($N$ models)} & 54.04 & 75.66 & 61.57 & 83.72 & 65.08 & 71.19 \\
    {BiLSTM-MultiCRF} & 54.38 & 74.42 & 60.23 & \textbf{84.70} & 67.29 & 73.97 \\
    {SUOpenTag} & 55.34 & 72.94 & 60.49 & 80.16 & 69.13 & 73.31 \\
    \textbf{AdaTag (Our Model)} & \textbf{56.05} & \textbf{76.07} & \textbf{62.00} & 82.90 & \textbf{75.48} & \textbf{78.45} \\
    \bottomrule 
\end{tabular}
}
\caption{Performance comparison on high-resource and low-resource attributes.}
\label{tab:high_low_resource}
\end{table}

\subsection{Ablation Studies}
\label{sec:ablation}

\paragraph{Attribute Embeddings.}
We study different choices of adopting pretrained attribute embeddings. 
Specially, we experiment with contextualized embeddings (BERT$_\text{name+value}$) and uncontextualized embeddings (Glove$_\text{name+value}$) under the ``Title'' setting.
For given attribute embeddings, we can either finetune them during training or freeze them once loaded.
We also experiment with attribute name embeddings $\mathbf{r}^\text{name}$ and attribute value embeddings $\mathbf{r}^\text{value}$ only to understand which information is more helpful.
The baseline is set as using randomly initialized learnable attribute embeddings.
Table~\ref{tab:att_emb} shows the results.
Comparing attribute embeddings with the same dimension, we find that freezing pretrained embeddings always leads to performance gain over the random baseline.
This is because our parameterization methods have high flexibility in generating the parameters for the decoder.
Using pretrained embeddings and freezing them provides the model with a good starting point and makes learning easier by reducing the degree of freedom.
BERT$_\text{name}$ (freeze) outperforms BERT$_\text{value}$ (freeze), suggesting that the attribute name is more informative in determining the characteristics of the value extraction task on our dataset, where the values labeled through distant supervision are noisy.

\begin{table}[ht]
\centering
\scalebox{0.7}{
\begin{tabular}{lcccc}
    \toprule
    \textbf{Attribute Embeddings} & \textbf{Dimension} & P(\%) & R(\%) & F$_1$(\%) \\
    \midrule
    {Random} & 100 & 63.05 & 72.35 & 64.82 \\
    {Glove$_\text{name+value}$} & 100 & 64.12 & 70.51 & 63.89 \\
    {Glove$_\text{name+value}$} (freeze) & 100 & \textbf{64.47} & \textbf{73.11} & \textbf{65.53} \\
    \midrule
    {Random} & 768 & 63.83 & 72.39 & 65.12 \\
    {BERT$_\text{name}$} & 768 & 62.01 & 73.94 & 64.89 \\
    {BERT$_\text{name}$} (freeze) & 768 & 64.90 & \textbf{74.31} & \textbf{66.60} \\
    {BERT$_\text{value}$} & 768 & \textbf{65.03} & 72.36 & 65.53 \\
    {BERT$_\text{value}$} (freeze) & 768 & 62.96 & 73.92 & 65.51  \\
    \midrule
    {Random} & 1536 & 64.80 & 71.95 & 65.74 \\
    {BERT$_\text{name+value}$} & 1536 & 63.57 & 73.57 & 65.81 \\
    {BERT$_\text{name+value}$} (freeze) & 1536 & \textbf{65.00} & \textbf{75.87} & \textbf{67.48} \\
    \bottomrule 
\end{tabular}
}
\caption{Performance (Title) with different choices for deriving and adopting attribute embeddings.}
\label{tab:att_emb}
% \vspace{-4mm}
\end{table}

\begin{table}[ht]
\centering
\scalebox{0.7}{
\begin{tabular}{ccccc}
    \toprule
    \textbf{Linear Layer} & \textbf{Transition Matrix} & P(\%) & R(\%) & F$_1$(\%) \\
    \midrule
    {MoE} & {MoE} & 42.28 & 65.80 & 47.94 \\
    {hypernetwork} & {hypernetwork} & \textbf{65.59} & 69.39 & 63.66 \\
    {MoE} & {hypernetwork} & 53.52 & 66.43 & 55.10 \\
    {hypernetwork} & {MoE} & 65.00 & \textbf{75.87} & \textbf{67.48} \\
    \bottomrule 
\end{tabular}
}
\caption{Performance (Title) with different parameterization methods.}
\label{tab:para}
\end{table}

\paragraph{Decoder Parameterization.}
We study different design choices for parameterizing the CRF-based decoder. For designs involving MoE, we search the number of experts ($k$) in $[1,2,3,4,5]$ and adopt the best one to present the results.
We experiment under the ``Title'' setting.
From Table~\ref{tab:para}, we find that parameterizing the linear layer with MoE leads to much lower performance.
This is reasonable because the linear layer plays a much more important role in the decoder while the transition matrix acts more like a regularization to avoid bad tag sequences.
MoE uses $k$ matrices as basis and expects to represent the parameters for any attribute as a linear combination of the bases.
That limits the expressiveness to capture complicated characteristics of different attributes and will thus severely hurt the performance.
As for the transition matrix, modeling with MoE is a better choice.
This is because the transition matrix is more ``structured'' in the sense that each of it element is expected to be either a big number or a small number based on its semantics.
For example, the transition score for $\text{I}\rightarrow\text{E}$ should be much higher than $\text{I}\rightarrow\text{B}$.
Hypernetwork is too flexible to generate such ``structured'' parameters.

% \subsubsection{Product-Level Evaluation}

% \begin{table}[ht]
% \centering
% \scalebox{0.7}{
% \begin{tabular}{ccccc}
%     \toprule
%     \textbf{Train} & \textbf{Inference} & P(\%) & R(\%) & F$_1$(\%) \\
%     \midrule
%     {Title} & {Title} &  &  &  \\
%     {Title} & {Title+Bullets} &  &  &  \\
%     {Title+Bullets} & {Title} &  &  &  \\
%     {Title+Bullets} & {Title+Bullets} &  &  &  \\
%     \bottomrule
% \end{tabular}
% }
% \caption{\textbf{Performance with product-level evaluation.}}
% \label{tab:product}
% \end{table}

\vspace{0mm}
\subsection{Effect of Number of Attributes}

% \begin{figure}[ht]
% \centering
% \includegraphics[scale=0.35]{figures/curve.png}
% \vspace{-0.3cm}
% \caption{Performance (Title) with different numbers of training attributes. \xiang{switch to bar chart will look better.}}
% \label{fig:scalability}
% \end{figure}

\begin{figure}[t]
\centering
\includegraphics[scale=0.14]{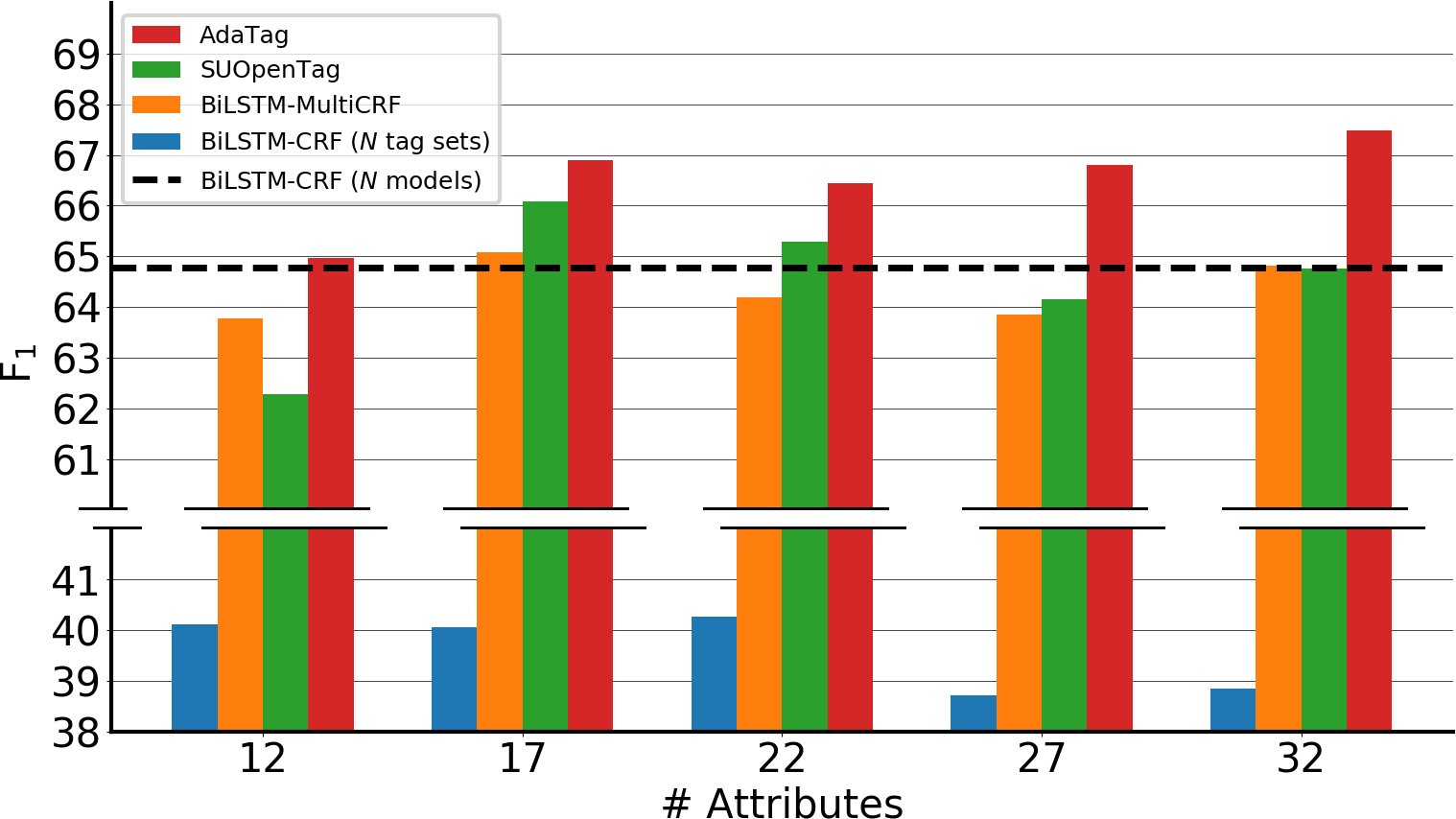}
% \vspace{-0.3cm}
\caption{Performance (Title) with different numbers of training attributes. We use broken y-axis due to the large gap in results between BiLSTM-CRF ($N$ tag sets) and other models.}
%The $#$ for BiLSTM-CRF ($N$ tag sets) is }
\label{fig:scalability}
% \vspace{-0.1cm}
\end{figure}

An important motivation of our model is that joint modeling of different attributes can facilitate knowledge sharing and improve the performance.
Here we study the performance of model improvement along with increment of the number of jointly modeled attributes.
We experiment under the ``Title'' setting.
We start with training our model on 12 attributes that have test data.
After that, we random select 5, 10, 15, 20 attributes from the remaining attributes, and add them to the joint training.
The evaluation results on 12 test attributes are presented in Figure~\ref{fig:scalability}. While our model general demonstrates greater improvement with joint modeling of more attributes, other models' performance fluctuate or goes down.
%which show that joint modeling of more attributes leads to greater improvement.
That also demonstrates the scalability of our model when new attributes keep emerging in real-world scenarios.

% \subsubsection{Case Study}
% \xiang{can have a small figure showing couple example output for comparison between 2-3 systems?}
\section{Related Work}
\paragraph{Attribute Value Extraction.}
OpenTag \citep{zheng2018opentag} formulates attribute value extraction as a sequence tagging task, and proposes a BiLSTM-SelfAttention-CRF architecture to address the problem.
\citet{xu2019scaling} propose an ``attribute-aware'' setup, by utilizing one set of BIO tags and attribute name embedding with an attention mechanism, to enforce the extraction network to be attribute comprehensive.
% \citet{wang2020learning} use both character-level and word-level embedding to alleviate the OOV issue in biomedical text. They also adopt a multi-task mechanism by training models with different entity types and shares parameters across these models.
\citet{karamanolakis2020txtract} additionally incorporate the product taxonomy into a multi-task learning setup, to capture the nuances across different product types. 
\citet{zhu2020multimodal} introduce a multi-modal network to combine text and visual information with a cross-modality attention to leverage image rich information that is not conveyed in text.
\citet{wang2020learning} use a question answering formulation to tackle attribute value extraction.
We adopt the extraction setup in our model as most of previous contributions, using sequence labeling architecture. But we utilize an adaptive decoding approach, where the decoding network is parameterized with the attribute embedding.

\paragraph{Dynamic Parameter Generation.}
% \xiang{this and next paragraphs seem overly specific to be discussed in related work. Can merge and compress into one paragraph for ``Dynamic parameter generation" or sth similar.}
Our model proposes an adaptive-based decoding setup, parameterized with attribute embeddings through a Mixture-of-Experts module and a hypernetwork.
% \paragraph{Mixture-of-Experts.}
\citet{jacobs1991adaptive} first propose a system composed of several different ``expert'' networks and use a gating network that decides how to assign different training instances to different ``experts''.
% In visual reasoning, \citet{jo2018modularity} adopt MoE with each expert being a stack of residual modules to achieve network modularity. In visual question answering, \citet{pahuja2019learning} use modularized ResNeXt-101 and treat each path as a ``expert'', and using gate controller to determine the most important path.
\citet{alshaikh2020mixture, guo2018multi, le2016lstm, peng2019text} all use domain/knowledge experts, and combine the predictions of each expert with a gating network.
Unlike these works, we combine the weights of each expert to parameterize a network layer given an input embedding.
\citet{ha2016hypernetworks} propose the general idea of generating the parameters of a network by another network. 
% \citet{noh2016image} tackle the VQA problem by learning a CNN with a dynamic parameter layer whose weights are controlled adaptively by the individual question. 
The proposed model in \citet{cai2020adaptive} generates the parameters of an encoder-decoder architecture by referring to the context-aware and topic-aware input.
\citet{suarez2017language} uses a hypernetwork to scale the weights of the main recurrent network.
\citet{platanios2018contextual} tackle neural machine translation between multiple languages using a universal model with a contextual parameter generator.

\section{Conclusion}
In this work we propose a multi-attribute value extraction model that performs joint modeling of many attributes using an adaptive CRF-based decoder.
Our model has a high capacity to derive attribute-specific network parameters while facilitating knowledge sharing.
Incorporated with pretrained attribute embeddings, our model shows marked improvements over previous methods.
%In the future, we plan to work on grouping attributes into clusters, where joint modeling of attributes in a cluster can maximize the performance gain. %(2) How to make it possible to get value predictions for all applicable attributes within one query.
\section*{Acknowledgments}
This work has been supported in part by NSF SMA 18-29268.
We would like to thank Jun Ma, Chenwei Zhang, Colin Lockard, Pascual Martínez-Gómez, Binxuan Huang from Amazon, and all the collaborators in USC INK research lab, for their constructive feedback on the work.
We would also like to thank the anonymous reviewers for their valuable comments.

\bibliographystyle{acl_natbib}
\bibliography{acl2021}

\clearpage
\appendix
\section{Number of Model Parameters}
\label{sec:num_parameters}

\begin{table}[ht]
\centering
\scalebox{0.7}{
\begin{tabular}{lr}
    \toprule
    \textbf{Methods} & \textbf{\# Parameters} \\
    \midrule
    {BiLSTM ($\mathbf{N}$ tag sets)} & {$0.6k\cdot \mathbf{N} + 6M$} \\
    {BiLSTM-CRF ($\mathbf{N}$ tag sets)} & {$9\cdot \mathbf{N}^2 + 0.6k\cdot \mathbf{N} + 6M$} \\
    {BiLSTM/BiLSTM-CRF ($\mathbf{N}$ models)} & {$6M\cdot \mathbf{N}$} \\
    {BiLSTM-CRF-SharedEmb} & {$0.1M\cdot \mathbf{N} + 6M$} \\
    {BiLSTM-MultiCRF} & {$2k\cdot \mathbf{N} + 6M$} \\
    {AdaTag} & {$8M$} \\
    \bottomrule 
\end{tabular}
}
\caption{Numbers of parameters for BiLSTM-based models with $\mathbf{N}$ attributes.}
\label{tab:num_parameters}
\end{table}

In our main experiment (Table~\ref{tab:main}), the numbers of parameters ($M=1,000,000$; $k=1,000$) for BiLSTM-based models with $\mathbf{N}$ attributes are listed in Table~\ref{tab:num_parameters}.
BERT (\texttt{bert-base-cased}) itself has $110M$ parameters, making BERT-based models generally much larger.

For our AdaTag, the weights for the hypernetwork ($\mathbf{W}_\text{hyper}^w\in\mathbb{R}^{4d_h\times d_r}$) have $(4\times 200)\times 1536$ parameters.
The number can be reduced by inserting a middle layer with fewer neurons.

\end{document}